\documentclass{article}

\usepackage{arxiv}
\usepackage{amsmath}

\usepackage[utf8]{inputenc} 
\usepackage[T1]{fontenc}    
\usepackage{hyperref}       
\usepackage{url}            
\usepackage{booktabs}       
\usepackage{amsfonts}       
\usepackage{nicefrac}       
\usepackage{microtype}      
\usepackage{lipsum}
\usepackage{graphicx}
\usepackage{array}
\title{Toward Efficient Breast Cancer Diagnosis and Survival Prediction Using L-Perceptron}

\author{
  Hadi Mansourifar\\
  Department of Computer Science\\
  University of Houston\\
  Houston, Texas \\
  \texttt{hmansourifar@uh.edu} \\
   \And
 Weidong Shi \\
  Department of Computer Science\\
  University of Houston\\
  Houston, Texas \\
  \texttt{wshi3@central.uh.edu} \\
}
\begin{document}
\maketitle

\begin{abstract}
Breast cancer is the most frequently reported cancer type among the women around the globe and beyond that it has the second highest female fatality rate among all cancer types.  Despite all the progresses made in prevention and early intervention, early prognosis and survival prediction rates are still unsatisfactory. In this paper, we propose a novel type of perceptron called L-Perceptron which outperforms all the previous supervised learning methods by reaching 97.42 \% and 98.73 \% in terms of accuracy and sensitivity, respectively in Wisconsin Breast Cancer dataset. Experimental results on Haberman's Breast Cancer Survival dataset, show the superiority of proposed method by reaching 75.18 \% and 83.86 \% in terms of accuracy and F1 score, respectively. The results are the best reported ones obtained in 10-fold cross validation in absence of any preprocessing or feature selection.
\end{abstract}


\section{Introduction}
Nowadays, cancer is the second leading cause of death in the U.S based on IARC report and it would replace heart disease as the main cause of death \cite{19}. Breast cancer is considered as one of the deadliest diseases with a high fatality rate among women worldwide. Early prognosis is the most effective way to minimize the physical and psychological side effects of prolonged treatments. Machine Learning methods are effective ways for early diagnosis and survival prediction \cite{8}. Over the past decades, large amounts of cancer data have been collected and are available to the Machine Learning and Data Mining community. In parallel, a wide range of supervised learning methods have been applied on collected datasets \cite{16,13,7,2,12}. However, due to vital impacts of correct diagnosis in cancer treatments, the accurate diagnosis and efficient survival prediction is still one of the most challenging tasks for researchers. In this paper, we propose a novel type of perceptron called L-Perceptron which outperforms all the previous supervised learning methods by reaching 97.42 \% and 98.73 \% in terms of accuracy and sensitivity, respectively in Wisconsin Breast Cancer dataset. Experimental results on Haberman's Cancer Survival dataset, show the superiority of proposed method by reaching 75.18 \% and 83.86 \% in terms of accuracy and F1 score, respectively. L-Perceptron has been devised by combination of least square classifiers and traditional perceptron   ideas. In L-perceptron, given $p1$ and $p2$ as $Y$ values, a mathematical function is fitted per feature same as least squares classification. What’s trained during update rule is characteristic of each function per feature. For example, in case of polynomials, the best possible polynomial order is trained per feature. This procedure helps L-Perceptron to be not only non-linear but very flexible to handle overfitting problem. The rest of this paper is organized as follows. Section 2 reviews some of the most important researches in breast cancer diagnosis and survival prediction. In section 3, we propose a novel type of perceptron called L-Perceptron. Section 4 reports experimental results on Wisconsin Breast Cancer and Haberman's Breast Cancer Survival datasets. Finally, section 5 concludes the paper.

\section{Related Work}
In this section, we provide a review on several studies which have been conducted on breast cancer diagnosis and survival prediction. These studies have focused on different approaches to the given problem and achieved high classification accuracies. \cite{18} proposed an interactive evaluation - diagnosis computer system based on cytologic features. Vikas Chaurasia and Saurabh Pal \cite{5} compared the performance criterion of supervised learning classifiers, such as Naive Bayes, SVM-RBF kernel, RBF neural networks, Decision Tree (DT) (J48), and simple classification and regression tree (CART), to find the best classifier in breast cancer datasets. The experimental result shows that SVM-RBF kernel is more accurate than other classifiers since it scores at the accuracy level of 96.84 \% in the Wisconsin Breast Cancer (original) dataset. Asri et al. \cite{1} compared the performance of C4.5, Naive Bayes, Support Vector Machine (SVM) and K- Nearest Neighbor (K-NN) to find the best classifier in Wisconsin Breast Cancer (original) showing that SVM proves to be the most accurate classifier with accuracy of 97.13 \%.  Vikas Chaurasia and Saurabh Pal \cite{6} used three popular data mining algorithms (Naive Bayes, RBF Network, J48) to develop the prediction models using the Wisconsin Breast Cancer (original). The obtained results indicated that the Naive Bayes performed the best with a classification accuracy of 97.36 \% and RBF Network came out to be the second best with a classification accuracy of 96.77 \%, and the J48 came out to be the third with a classification accuracy of 93.41 \%. \cite{11} used an evolutionary artificial neural network (EANN) approach based on the pareto-differential evolution (PDE) algorithm augmented with local search for the prediction of breast cancer. The approach is named Memetic Pareto Artificial Neural Network (MPANN).  Since the early dates of the researches in the field of computational biomedicine, the cancer survivability prediction has been a challenging problem for many researchers \cite{9,3}. In \cite{10}, artificial neural networks and decision trees along with logistic regression were used to develop the prediction models using 202,932 breast cancer patients records, which then pre-classified into two groups of “survived” (93,273) and “not survived” (109,659). The results of predicting the survivability were in the range of 93 \% accuracy.\cite{15} used a fully complex valued fast learning artificial neural network with GD activation function in the hidden layer. The comparison results showed that, FC-FLC provides a better classification performance comparing to the SRAN, MCFIS and ELM classifiers. Liu et al. \cite{14} used the under-sampling C5 technique and bagging algorithm to deal with the imbalanced problem predictive models for breast cancer survivability.
\section{Proposed Method}
In this section, we propose a novel type of perceptron called L-Perceptron which despite its simplicity, it can outperform traditional supervised learning methods in breast cancer diagnosis and survival prediction.

\subsection{L-Perceptron}
Given a set of $n$ training data in $m$ dimensional space $X= \{x_1, x_2, x_3..., x_n\}$, a set of corresponding labels  $Y=\{y_1,y_2,y_3,...,y_n \}$, $p1$ and $p2$ as two hyperparameters, $V=\{v_1,v_2,v_3,...,v_n\}$ is created as follows. 
\[ \begin{cases} 
      p1    & if \quad y_ {j}=class 1 \quad (positive \quad instance) \\
     p2  &  otherwise 
      
   \end{cases}
\]
Where, $p1$ and $p2$ can be manipulated by the user to get the best possible result. Training phase of a L-Perceptron starts by fitting function $F$ by minimizing $S$, given $V$ per each feature as follows.
\[ r_i= [v_i -F (x_i, \alpha)]\] 
\[ S= \sum_{i=1} ^{n} (r_i)^2 \]
After finding the best fitting function per each feature test phase is formulized as follows.
\[ \begin{cases} 
      1    & if \sum_{j=1} ^{m} F_j(X) > 0 \\
     0  &  otherwise 
      
   \end{cases}
\]
Where, $F_j$ is fitting function of $j^{th}$ feature. It means that instead of dot product of weights and features the features are passed to their corresponding function and a summation of functions outputs is passed to a step function or activation function to predict the label of input test data. The diagram of L-Perceptron is shown in figure1.Before starting the training phase, the fitting function type must be defined which can be selected among all possible mathematical functions like logarithmic, exponential, polynomial, etc. What’s trained during the update rule is characteristic of the fitting function. For example, in case of polynomials the best possible polynomial order is trained per feature. This lets the model to meet each feature set complexity separately.
\begin{figure}[ht!]
\centering
\includegraphics[width=90mm]{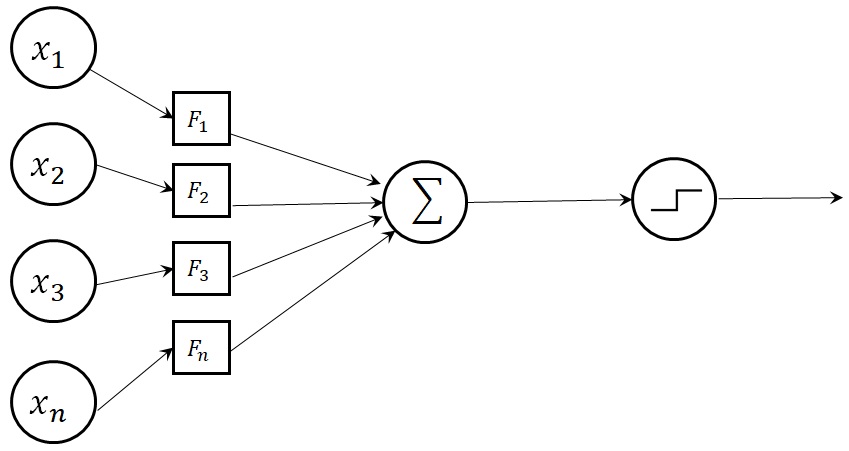}
\caption{Schematic of L-Perceptron. \label{overflow}}
\end{figure}

\subsection{Update Rule}
Suppose polynomials are used as fitting function. In this case, the order of each fitting polynomial is trained during the update rule. What’s happening during the update rule is to assign initial polynomial orders, fit a polynomial per feature given the $p1$ and $p2$, calculate the error and repeat this procedure until the best possible orders are found at the end of update rule. To make the update rule faster the upper bound and lower bound of polynomial orders can be limited to some specific range. The upper bound and lower bound range can be defined as hyperparameters as described in implementation section. This restriction adds more flexibility to the L-Perceptron to avoid overfitting. Update rule of L-Perceptron is as follows.

\rule[1mm]{125mm}{0.5mm}
\\ 
\textbf{1.}	Initialize all degrees to 1 and $Error=a$ \\
\textbf{2.}	Repeat until no significant change seen in $Error$ \\
\textbf{3.}	\quad \quad \quad          For all Features \\
\textbf{4.}	 \quad \quad \quad \quad \quad           $degree[i]=degree[i]+1$ \\
\textbf{5.}	  \quad \quad \quad \quad \quad                Compute the $NewError$ \\
\textbf{6.}	               \quad \quad \quad \quad \quad    if ($NewError< Error$){   $Error = NewError$} \\
\textbf{7.}	               \quad \quad \quad \quad \quad  else{ $degree[i]=degree[i]-1$} \\
\rule[0mm]{125mm}{0.5mm}\\
\begin{figure}[ht!]
\centering
\includegraphics[width=90mm]{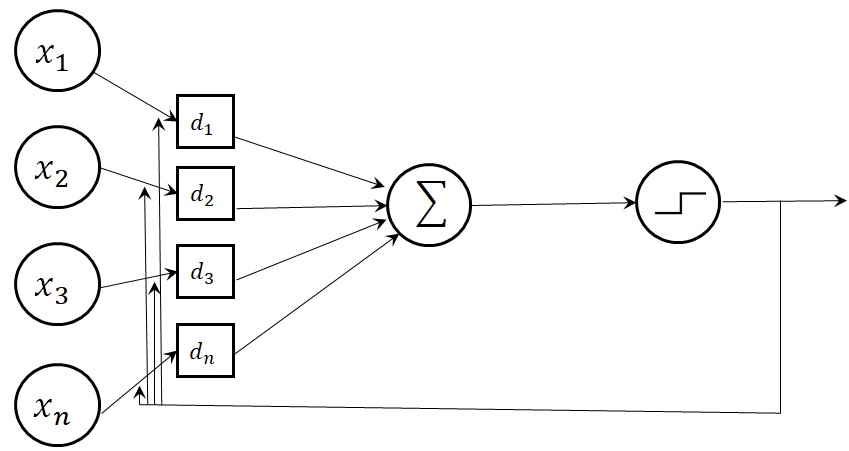}
\caption{Schematic of L-Perceptron training phase. \label{overflow}}
\end{figure}

Figure 2 shows the schematic of L-Perceptron, in case of using polynomials as fitting function. 
\section{Experiments}
In this section, we compare the accuracy results of proposed method versus a set of traditional supervised learning methods by testing on WBCD and HSD. 
\subsection{Datasets}
In this section, we provide a brief introduction to datasets used in the experiments including Wisconsin Breast Cancer dataset (original) and Haberman's Breast Cancer Survival dataset.
\subsubsection{Wisconsin Breast Cancer dataset}
The breast cancer dataset is a classic binary classification dataset. The data is accessible from the UC Irvine Machine Learning repository \cite{17}. This dataset has 699 instances, two classes (malignant and benign), and 9 integer-valued attributes. It contains 16 instances with missing values, but we didn’t remove them from the dataset. It’s consisted of benign: 458 (65.5\%), malignant: 241 (34.5\%) instances. 

\begin{table}[!htb]
\centering
\begin{tabular}{|c|c|}
\hline
            & \textbf{Attribute}          \\ \hline
\textbf{1}  & Sample code number          \\ \hline
\textbf{2}  & Clump thickness             \\ \hline
\textbf{3}  & Uniformity of cell size     \\ \hline
\textbf{4}  & Uniformity of cell shape    \\ \hline
\textbf{5}  & Marginal adhesion           \\ \hline
\textbf{6}  & Single epithelial cell size \\ \hline
\textbf{7}  & Bare nuclei                 \\ \hline
\textbf{8}  & Bland chromatin             \\ \hline
\textbf{9}  & Normal nucleoli             \\ \hline
\textbf{10} & Mitoses                     \\ \hline
\textbf{11} & Class                       \\ \hline
\end{tabular}
\end{table}
\subsubsection{ Haberman's Breast Cancer Survival dataset}
The Haberman's Breast Cancer Survival dataset \cite{4} contains of cases from a study that was conducted between 1958 and 1970 at the University of Chicago's Billings Hospital on the survival of patients who had undergone surgery for breast cancer. This dataset has 306 instances, 2 classes and 3 integer-valued attributes. Each data point contains following features.
\begin{table}[!htb]
\centering
\begin{tabular}{|l|l|}
\hline
           & \textbf{Attribute}                         \\ \hline
\textbf{1} & Age of patient at time of operation        \\ \hline
\textbf{2} & Patient's year of operation (year - 1900)  \\ \hline
\textbf{3} & Number of positive axillary nodes detected \\ \hline
\textbf{4} & Class                                      \\ \hline
\end{tabular}
\end{table}
Output attribute is a survival status (class attribute) assigned for the output attribute is as follows:
 Patient lived for 5 years or longer =1.
 Patient death within 5 years =2.
\subsubsection{Implementation}
We have developed a Python function called “lp.py” which is available for the public to test on classification problems \cite{20}. This function has been originally devised for numerical datasets. However, it can be used on categorical datasets if the features supposed as a set of discrete integer values. The input parameters of this functions are as follows.\\
\textbf{lp}(\textit{trainx}, \textit{trainy}, \textit{testx}, \textit{testy}, \textit{p1}, \textit{p2}, \textit{dlb}, \textit{dub}, \textit{ite}, \textit{threshold})\\
Table1 shows implemented parameters and their descriptions.
\begin{table}[!htb]
\centering
\begin{tabular}{|l|l|}
\hline
\textbf{Parameter}      & \textbf{Description}                                                              \\ \hline
\textbf{trainx, trainy} & A set of training data and their corresponding labels                             \\ \hline
\textbf{testx, testy}   & A set of test data and their corresponding labels                                 \\ \hline
\textbf{p1, p2}         & Y values for fitting functions as described in section 3                          \\ \hline
\textbf{dlb, dub}       & Degree lower bound and degree upper bounds respectively to limit the degree range \\ \hline
\textbf{ite}            & Number of iterations                                                              \\ \hline
\textbf{threshold}      & Threshold of activation function to discriminate between the instances.           \\ \hline
\end{tabular}
\bigskip
\caption{  Parameters and their description.}
\end{table}

\subsubsection{Results and discussion }
We used 10-fold cross validation method to measure the unbiased estimation for performance comparison purposes. The comparison results of different methods tested on Wisconsin Breast Cancer Dataset (WBCD) are presented in Table 2. Experimental results on Haberman’s Survival Dataset (HSD) have been tabulated on Table 3. Table 2 shows that L-Perceptron outperforms other methods in terms of accuracy and sensitivity based on experiments on WBCD. Naive Bayes is in the second rank by showing better results comparing to the rest of methods.
\begin{table}[htb!]
\centering
\begin{tabular}{|c|c|c|c|c|}
\hline
\textbf{Methods} & \textbf{Accuracy (\%)} & \textbf{Sensitivity (\%)} & \textbf{Specificity (\%)} & \textbf{F1 Score (\%)} \\ \hline
L-Perceptron     & 97.42                  & 98.73                      & 96.2                     & 96.50                  \\ \hline
Naive Bayes      & 97.36                  & 97.4                      & 97.9                      & 97.64                  \\ \hline
RBF Network      & 96.77                  & 97.07                     & 96.23                     & 96.6                   \\ \hline
J48              & 93.41                  & 93.4                      & 90.37                     & 91.86                  \\ \hline
\end{tabular}
\bigskip
\caption{Accuracy, sensitivity, specificity and F1 score of different methods tested on Wisconsin Breast Cancer Dataset.}
\end{table}
\begin{figure}[htb!]
\centering
\includegraphics[width=120mm]{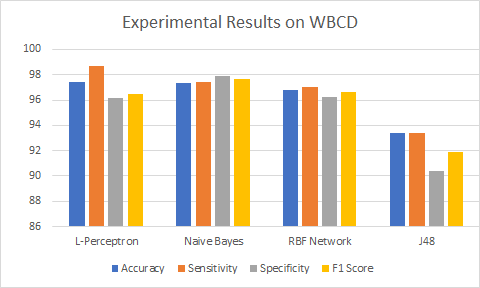}
\caption{Comparative graph of different classifiers tested on WBCD. \label{overflow}}
\end{figure}
\begin{table}[htb!]
\centering
\begin{tabular}{|c|c|c|c|c|}
\hline
\textbf{Methods}                      & \textbf{Accuracy (\%)} & \textbf{Sensitivity (\%)} & \textbf{Specificity (\%)} & \textbf{F1 Score (\%)} \\ \hline
\textbf{L-Perceptron}                 & 75.18                  & 90.04                     & 37.08                     & 83.86                  \\ \hline
\textbf{Logistic Regression}          & 74.27                  & 94.77                     & 22.95                     & 82.62                  \\ \hline
\textbf{Linear Discriminant Analysis} & 73.78                  & 95.42                     & 19.67                     & 82.71                  \\ \hline
\textbf{KNN}                          & 71.03                  & 88.23                     & 34.42                     & 81.57                  \\ \hline
\textbf{CART}                         & 64.02                  & 74.5                      & 26.22                     & 78.44                  \\ \hline
\textbf{Naive Bayes}                  & 74.17                  & 94.11                     & 27.86                     & 82.52                  \\ \hline
\textbf{SVM}                          & 69.77                  & 95.42                     & 3.27                      & 82.71                  \\ \hline
\textbf{MLP}                          & 66.21                  & 62.74                     & 55.73                     & 72.64                  \\ \hline
\textbf{Random Forest}                & 67.27                  & 81.69                     & 22.95                     & 80.38                  \\ \hline
\end{tabular}
\bigskip
\caption{Accuracy, sensitivity, specificity and F1 score of different methods tested on Haberman’s Survival Dataset.}
\end{table}

Table 3 shows that L-Perceptron outperforms other methods in terms of accuracy and F1 score based on experiments on HSD. When it comes to compare the methods based on specificity, L-Perceptron is at the second rank after MLP which shows best specificity result among all tested methods. Table 4 shows the initialized parameters including $p1$, $p2$ and fitting degree for each dataset. \\
\\
\\
\\
\begin{figure}[ht!]
\centering
\includegraphics[width=170mm]{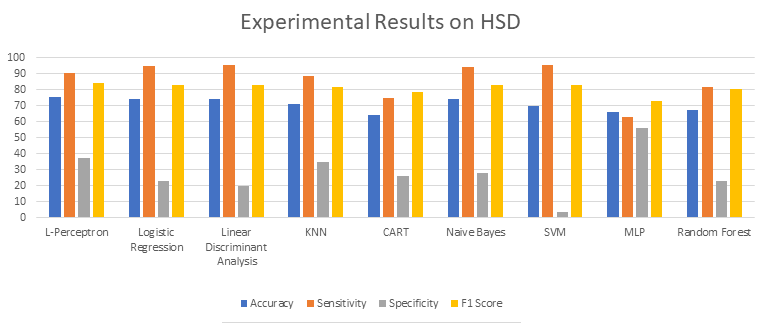}
\caption{Comparative graph of different classifiers tested on HSD. \label{overflow}}
\end{figure}
\begin{table}[htb!]
\centering
\begin{tabular}{|c|c|c|}

\hline
\textbf{Parameter}          & \textbf{WBCD} & \textbf{HSD} \\ \hline
\textit{\textbf{p1, p2}}    & -2, 3         & -1.3, 2.9    \\ \hline
\textit{\textbf{dlb, dub}}  & 4, 4          & 1, 1         \\ \hline
\textit{\textbf{ite}}       & 2             & 0            \\ \hline
\textit{\textbf{threshold}} & 0.5           & 0.42         \\ \hline
\end{tabular}
\bigskip
\caption{. Initialized parameters used in experiments.}
\end{table}
\section{Conclusion}
In this paper, we proposed a novel type of perceptron called L-Perceptron. The proposed method successfully tested on Wisconsin Breast Cancer Dataset and Haberman’s Breast Cancer Survival Dataset. We used 10-fold cross-validation method to measure the unbiased estimation for performance comparison purposes. The experimental results showed that L-Perceptron has the best performance comparing to previous methods in terms of accuracy and sensitivity based on experiments on WBCD. The proposed method reached 97.42 \% of accuracy, 98.73 \% of sensitivity which are the best performance results reported in the literature among the reported results without any preprocessing or feature selection. We also tested L-Perceptron on Haberman’s Breast Cancer Survival Dataset. The experimental results showed that L-Perceptron has the best performance comparing to previous methods in terms of accuracy and F1 score. The proposed method reached 75.18 \% of accuracy, 83.86 \%  of F1 score which are the best reported performance results.
\bibliographystyle{unsrt}

\end{document}